\documentclass[10pt,letterpaper]{article}

\usepackage[letterpaper,margin=1in]{geometry}
\usepackage[T1]{fontenc}
\usepackage[utf8]{inputenc}
\usepackage{times}
\usepackage{microtype}
\usepackage{amsmath,amssymb,amsfonts}
\usepackage{amsthm}
\usepackage{mathrsfs}
\usepackage{graphicx}
\graphicspath{{./}{Preprint/}}
\usepackage{multirow}
\usepackage{makecell}
\usepackage[caption=false,font=normalsize,labelfont=sf,textfont=sf]{subfig}
\usepackage{stfloats}
\usepackage{url}
\usepackage{verbatim}
\usepackage[dvipsnames,table]{xcolor}
\usepackage{xurl}
\usepackage{tabularx}
\usepackage{booktabs}
\usepackage{adjustbox}
\usepackage[numbers,sort&compress]{natbib}
\usepackage{algorithm}
\usepackage{algpseudocode}
\usepackage{algorithmicx}
\usepackage{listings}
\usepackage{fancyhdr}
\usepackage{titlesec}
\usepackage[section]{placeins}
\usepackage{hyperref}

\hypersetup{
    pdfpagemode=UseNone,
    colorlinks=true,
    linkcolor=blue,
    filecolor=magenta,  
    citecolor=blue,    
    urlcolor=blue,
    pdftitle={BLPR: Robust License Plate Recognition under Viewpoint and Illumination Variations via Confidence-Driven VLM Fallback},
    pdfauthor={Guillermo J. Auza Banegas, Diego A. Calvimontes Vera, Natalia L. Condori Peredo, Edwin Salcedo, Sergio S. Castro Sandoval}
}
\definecolor{YesBlue}{RGB}{59, 76, 192}
\definecolor{NoRed}{RGB}{180, 4, 38}

\theoremstyle{plain}

\theoremstyle{remark}

\theoremstyle{definition}

\titleformat{\section}{\large\bfseries}{\thesection}{0.75em}{}
\titleformat{\subsection}{\normalsize\bfseries}{\thesubsection}{0.75em}{}
\titlespacing*{\section}{0pt}{2.0ex plus 0.5ex minus 0.2ex}{1.0ex}
\titlespacing*{\subsection}{0pt}{1.6ex plus 0.4ex minus 0.2ex}{0.8ex}

\fancypagestyle{firstpage}{
  \fancyhf{}
  \fancyfoot[L]{}

}

\begin{document}

\renewcommand{\thefootnote}{\fnsymbol{footnote}}
\thispagestyle{firstpage}
\begin{center}
{\LARGE\bfseries BLPR: Robust License Plate Recognition under Viewpoint and Illumination Variations via Confidence-Driven VLM Fallback\par}
\vskip 1.2em
{\normalsize
Guillermo J. Auza Banegas\textsuperscript{1},\quad
Diego A. Calvimontes Vera\textsuperscript{1},\quad
Natalia L. Condori Peredo\textsuperscript{1}\footnotemark[2],\\
Edwin Salcedo\textsuperscript{2}\footnotemark[1]\footnotemark[2],\quad
Sergio S. Castro Sandoval\textsuperscript{1}
}
\vskip 0.9em
{\small
\textsuperscript{1}Department of Mechatronics Engineering, Universidad Catolica Boliviana ``San Pablo'', La Paz, Bolivia\\
\textsuperscript{2}School of Electronic Engineering and Computer Science, Queen Mary University of London, England, United Kingdom\\
\href{https://github.com/EdwinTSalcedo/BLPR}{https://github.com/EdwinTSalcedo/BLPR}
}
\end{center}
\footnotetext[1]{Corresponding author: \href{mailto:e.r.salcedoaliaga@qmul.ac.uk}{e.r.salcedoaliaga@qmul.ac.uk}.}
\footnotetext[2]{These authors contributed equally to this work.}

\vskip 0.8em
\begin{center}
{\bfseries Abstract}
\end{center}
\begin{quote}
\small
Robust license plate recognition in unconstrained environments remains a significant challenge, particularly in underrepresented regions with limited data availability and unique visual characteristics, such as Bolivia. Recognition accuracy in real-world conditions is often degraded by illumination changes and viewpoint distortion. To address these challenges, we introduce BLPR, a deep learning-based License Plate Detection and Recognition (LPDR) framework designed for Bolivian license plates. BLPR adaptively applies geometric rectification, illumination correction, and VLM-assisted fallback based on image-condition and confidence cues. The proposed system uses a YOLO-based detector pretrained on synthetic data generated in Blender to simulate extreme perspectives and lighting conditions, and is fine-tuned on street-level data collected in La Paz, Bolivia. Detected plates are processed by a YOLO-based character recognizer, while a lightweight vision-language model (Gemma3 4B) is selectively triggered in ambiguous cases as a confidence-driven fallback mechanism. We also introduce the first publicly available Bolivian LPDR dataset for academic research, supporting evaluation under diverse viewpoint and illumination conditions. The system achieves a character-level recognition accuracy of 89.6\% on real-world data, demonstrating its effectiveness for deployment in challenging urban environments.
\end{quote}

{\small\noindent\textbf{Keywords:} Bolivian License Plate Recognition; Viewpoint Robustness; Illumination Robustness; Vision-Language Model Fallback; Bolivian License Plate Dataset\par}
\vskip 1.2em

\section{Introduction}
\label{sec:introduction}

Vehicle identification and registration in Latin America, particularly in Bolivia, continue to face persistent challenges due to the limited availability of robust technological tools for vehicle identification and verification. This limitation affects both citizens and authorities, contributing to issues such as car smuggling, duplicated license plates, and illegal transit. For example, in Bolivia, it is estimated that approximately 20\% of the national vehicle fleet consists of smuggled vehicles, many of which were reportedly stolen in neighboring countries \cite{reduno2025}. Official reports from the Bolivian Police indicate a substantial increase in reported property-related crimes, rising from 14,776 cases in 2020 to 21,330 cases in 2023, representing an increase of approximately 44.3\% over this period \cite{elpais2025}. This upward trend further underscores the need for reliable vehicle identification and monitoring systems. To the best of our knowledge, no publicly available dataset or license plate detection and recognition (LPDR) framework specifically targeting Bolivian license plates currently exists.

LPDR systems typically convert imagery captured by roadside or in-vehicle cameras into a structured identifier that can be queried in real-time for applications such as access control, tolling, traffic monitoring, and enforcement-oriented analytics. These systems generally involve four stages: vehicle image capture, license plate detection, character segmentation, and optical character recognition (OCR) \cite{patel2013anpr}. In addition to these stages, vehicle detection and tracking are often incorporated at the beginning of LPDR pipelines to improve system robustness. Since their development, numerous studies have focused on rectification techniques to correct skewed plates, thereby facilitating the character recognition stage \cite{salminen2023anpr, wang2018}. Despite impressive benchmark progress, LPDR accuracy still degrades sharply under uncontrolled, real-world conditions, particularly due to the combined effects of viewpoint distortion (e.g., wide-angle capture, oblique approach angles, and plate tilt) and illumination variability (e.g., shadows, glare, low light, and weather-induced photometric changes), where failures in plate localization propagate to character recognition \cite{laroca2018}.

To address these challenges, this paper presents BLPR, a novel LPDR framework for Bolivian plates that preserves real-time practicality while improving performance under viewpoint and illumination variability. While end-to-end models require training on large-scale datasets to generalize well to new domains \cite{lin2023}, modular pipelines enable the reuse of existing components and facilitate targeted training and customization of specific modules, such as OCR, license plate geometric normalization, and targeted pre-processing, using smaller datasets. The present work leverages a modular framework and synthetic data to mitigate annotation scarcity and improve coverage of rare viewpoints and lighting. It also incorporates a vision-language model as a fallback mechanism for plates with low recognition confidence and evaluates its effectiveness in the Bolivian context.

The main contributions are:

\begin{itemize}
\item The first publicly available Bolivian License Plate Recognition (BLPR) dataset, comprising real street scenes and synthetic renders, with annotations for detection and recognition and explicit categorization by viewpoint and illumination conditions.
\item A YOLO-based vehicle/plate localization stage with spatial validation, followed by illumination-aware pre-processing and perspective rectification to normalize plate geometry before recognition.
\item A fast character recognizer supported by a locally hosted vision-language model that is triggered only in low-confidence cases and constrained by Bolivian plate syntax to resolve ambiguous reads.
\item An ablation-driven robustness evaluation reporting accuracy, error rates, and runtime across viewpoint and illumination conditions, enabling transparent comparison of preprocessing, rectification, OCR, and fallback strategies.
\end{itemize}

The remainder of this paper reviews related LPDR approaches and robustness strategies, then details the proposed dataset, pipeline, and evaluation methodology.

\section{Background and Related Work}
\label{sec:related-works}

License plate detection and recognition (LPDR) systems can be broadly classified into two categories: end-to-end models and multi-stage pipelines. On the one hand, end-to-end approaches aim to directly map raw images to license plate text using a single trainable model, often combining plate detection and character recognition within a unified architecture. On the other hand, modular pipelines decompose the task into distinct stages, typically including plate detection, geometric normalization, character recognition, and post-processing. While end-to-end models are increasingly popular and computationally efficient, their performance strongly depends on the availability of large and diverse training datasets, and may struggle to generalize in low-data scenarios \cite{Xu2018EndToEndLPDR}. This trade-off motivates the design of the proposed system, which adopts a modular approach. This section reviews the most relevant prior work.

\subsection{Modular License Plate Recognition}

Typically, LPDR systems begin with license plate localization in an image using either handcrafted visual features (e.g., edges and contours) or data-driven detection models. The regions of interest (ROIs) are then refined through geometric rectification and photometric enhancement (e.g., noise reduction and contrast adjustment). Next, the refined image is segmented into individual character regions, commonly using projection-based methods. Finally, each character is classified using techniques ranging from template matching and support vector machines (SVMs) to modern deep learning (DL) models \cite{Yang2022LPDR}. This modular separation of license plate detection, character segmentation, and recognition has dominated LPDR research because it simplifies optimization and error isolation \cite{Xu2018EndToEndLPDR}, particularly important in low-data contexts.

Within this modular framework, optical character recognition (OCR) refers specifically to the process of extracting and recognizing the alphanumeric characters from a previously detected license plate region. Traditional OCR approaches commonly rely on morphological operations and filtering techniques for character segmentation, often combined with template matching methods \cite{Suryanarayana2005Morphology, Lin2019Morphology}. These pipelines typically include pre-processing steps such as Gaussian filtering, adaptive histogram equalization, edge detection, and morphological transformations to enhance character regions and isolate individual symbols. In recent years, DL techniques have been introduced to improve robustness and recognition accuracy; for example, \cite{Abtahi2015DeepRLSegmentation} employed deep reinforcement learning for character segmentation. Furthermore, several studies propose hybrid approaches that combine classical image processing techniques for character extraction with DL-based models, such as convolutional neural networks (CNNs) or lightweight OCR modules, for character recognition \cite{Lokesh2023HybridILPR}. The advantages of these hybrid frameworks, which leverage the efficiency of traditional pre-processing while benefiting from the feature learning capabilities of modern DL methods, motivated the design of the BLPR system.

\subsection{Robustness to Viewpoint Changes and Illumination Variations} 

License plates captured at tolls or by roadside and in-vehicle cameras often exhibit significant character warping, which severely degrades recognition performance and causes false positives. To improve character readability and correct distortions caused by the camera viewpoint, license plate rectification is commonly applied. Traditional geometric rectification approaches typically relied on affine transformations or Hough-based line estimation to recover a frontal view \cite{Silva2018LPR-Unconstrained}. Moreover, traditional approaches relied on segmentation of the rectified plate to isolate individual characters, which were then compared against predefined templates \cite{Lin2019Morphology, Yogheedha2018Template-Matching}. More recently, DL-based approaches enhance robustness to viewpoint changes by using instance segmentation to precisely delineate license plate regions and infer their geometric structure, facilitating subsequent perspective correction \cite{lin2023}.

Another challenge affecting most LPDR systems deployed in unconstrained scenarios is illumination variability, including reflective glare from the vehicle's plate, varying weather conditions, headlight glare, and changes in chassis appearance under different lighting conditions. To address these issues, numerous approaches have been developed over the years. Early efforts relied on traditional image processing techniques, such as using the HSV color space to segment saturation components and mitigate noise in the saturation channel \cite{Feng2010HSV}. More recent approaches, however, leverage DL methods. For instance, \cite{Boby2022GANN} implement generative adversarial networks (GANs) to enhance overall image quality prior to recognition. Similarly, \cite{Sultan2023HSV} propose a hybrid framework integrating Faster R-CNN for vehicle detection, morphological operations for plate localization, and a deep neural network (DNN) for recognition. Their approach incorporates HSV-based color segmentation and morphological processing to enhance plate regions and improve robustness under challenging conditions such as illumination variations and color inconsistencies. The method is evaluated on multiple challenging datasets, including PKU \cite{yuan2016license}, AOLP \cite{hsu2012application}, and CCPD \cite{Xu2018EndToEndLPDR}.

Despite these methodological advances, it is important to note that much of the reported success in LPDR has been achieved under relatively controlled or ideal conditions. Although high performance metrics have been demonstrated in terms of detection and recognition, model robustness in unconstrained and adverse environments remains an open research challenge. Several studies attempt to address this limitation by targeting specific environmental conditions. For example, \cite{Liu2024Foggy} focuses on improving model performance in foggy weather scenarios, while \cite{Rio-Alvarez2019Weather-Ilumination} adopts a broader approach to handle various weather-related adversities, including rainfall and illumination changes. These efforts highlight that achieving robustness to illumination variation is closely intertwined with the ability to generalize across diverse real-world conditions.

More recently, LPDR systems increasingly leverage publicly available datasets for training and validation in tasks such as plate detection, rectification, and OCR (see Table \ref{tab:lpr-regional-datasets}). Nevertheless, these datasets lack sufficient perspective distortion and illumination variability to assess robustness to such changes. In the Bolivian context, no publicly available dataset targeting Bolivian license plates exists, resulting in limited context-specific data for this task. This highlights the need for a dedicated dataset tailored to this scenario.

\begin{table}[!htbp]
  \centering
  \caption{A comparison of publicly available license plate recognition datasets and the BLPR dataset.}
  \label{tab:lpr-regional-datasets}
  \begin{adjustbox}{max width=\textwidth}
  \begin{tabular}{@{}l c c c c c c c c l@{}}
    \toprule
    \textbf{Dataset} & \textbf{Samples} & 
    \makecell{\textbf{Perspective} \\ \textbf{Variation}} & 
    \makecell{\textbf{Illumination} \\ \textbf{Variation}} & 
    \makecell{\textbf{Text} \\ \textbf{Annotations}} & 
    \makecell{\textbf{Plate } \\ \textbf{Detection}  \\ \textbf{Annotations}} & 
    \makecell{\textbf{Character} \\ \textbf{Detection}  \\ \textbf{Annotations}} &
    \makecell{\textbf{Car} \\ \textbf{Detection}  \\ \textbf{Annotations}} &
    \textbf{Country} & \textbf{Year} \\
    \midrule
    CCPD 2019 \cite{Xu2018EndToEndLPDR} & 355,013 & yes & yes & yes & yes & no & no & China & 2019\\
    CBLPRD-330k\cite{cblprd330k2023} & 330,000 & no & no & yes & yes & no & no & China & 2023\\
    RodoSol-ALPR \cite{laroca2022} & 20,000 & no & no & yes & yes & no & no & Brazil & 2022\\
    CCPD 2020 \cite{Xu2018EndToEndLPDR} & 11,776 & no & no & yes & yes & no & no & China & 2020\\
    UFPR-ALPR \cite{UFPR-ALPR2018} & 4,500 & no & no & yes & yes & no & yes & Brazil & 2018\\
    Artificial Mercosur \cite{ArtificialMercosur2020} & 3,829 & no & no & no & yes & no & no & Brazil & 2020\\
    AOLP \cite{AOLPD2012} & 2,049 & no & no & yes & yes & yes & no & Taiwan & 2012 \\
    SSIG-SegPlate \cite{SegPlate2016} & 2,000 & no & no & yes & yes & no & no & Brazil & 2016\\
    \midrule
    \textbf{BLPR-A} & 3,541 & yes & yes & no & yes & no & yes & Bolivia & 2026\\
    \textbf{BLPR-B} & 5,292 & yes & yes & yes & yes & yes & no & Bolivia & 2026\\
    \textbf{BLPR-C} & 520 & no & no & yes & no & yes & no & Bolivia & 2026\\
    \textbf{BLPR-D} & 503 & yes & yes & yes & no & yes & no & Bolivia & 2026\\ 
    \bottomrule
  \end{tabular}
  \end{adjustbox}
\end{table}

\subsection{Synthetic Data Generation and Domain Adaptation for LPDR Systems}

The robustness of DL-based LPDR systems to domain shift at deployment is strongly influenced by the availability and representativeness of annotated data. In regions with limited data collection infrastructure, such as several countries in Latin America, large-scale and diverse datasets remain scarce. Consequently, synthetic data generation has emerged as an effective strategy for improving model generalization under varying imaging conditions, including illumination changes, viewpoint distortions, occlusions, and plate formatting differences. Classical approaches based on template rendering and character permutation have demonstrated measurable gains in LPDR performance \cite{ribeiro2019}.

Recent advances in generative modeling have further enhanced the realism of synthetic data. GAN-based approaches, such as AsymCycleGAN \cite{Zhang2020RobustLPR}, translate synthetic plates into more realistic images by introducing variations in lighting, shadows, and perspective. More recently, diffusion-based models have enabled controllable and high-fidelity plate generation under diverse conditions \cite{shpira2024}. Despite these improvements, most studies evaluate synthetic data within limited regional contexts or controlled benchmarks, leaving cross-domain generalization insufficiently explored.

Beyond visual realism, research has increasingly focused on reducing the synthetic-to-real domain gap as a deployment problem rather than a purely perceptual one. Cross-dataset evaluations show that within-dataset performance can substantially overestimate real-world generalization, with significant accuracy drops under leave-one-dataset-out protocols and evidence of dataset-specific biases that models may exploit \cite{laroca2022,laroca2022first}. Accordingly, modern pipelines emphasize data-centric strategies, combining large synthetic corpora with limited real data through multi-source synthesis and quality filtering, as well as semi-supervised self-training (e.g., pseudo-labeling) to leverage unlabeled or weakly labeled target-domain images \cite{laroca2025,shpira2024}. This highlights opportunities for novel techniques, such as synthetic data generation, domain adaptation, and principled evaluation, in LPDR systems intended for deployment in underrepresented regions, such as Latin America.

\subsection{VLM-Assisted OCR and Post-processing}

OCR for outdoor scenes, and specifically for LPDR systems, has evolved from CNN-based recognizers to transformer-based architectures and, more recently, VLMs, which formulate plate reading as conditional generation or multimodal question answering. VLMs, such as GPT-4V, Flamingo, and BLIP-2, incorporate stronger language priors and enable joint visual–textual reasoning. However, they remain constrained by high computational cost and latency, requiring substantially more inference time and resources than dedicated OCR or LPDR pipelines, thereby limiting their deployability in real-time, resource-constrained environments. Consequently, while VLMs offer a promising direction for unified perception and reasoning, they are not yet viable substitutes for specialized recognition systems in practical deployments.

In contrast, VLMs have been increasingly explored as auxiliary components to enhance traditional OCR pipelines. Rather than replacing visual recognition, these models can be used as fallback mechanisms to improve predictions, particularly in low-confidence or ambiguous cases. Conventional OCR engines (e.g., Tesseract and Kraken) often rely heavily on task-specific pre-processing pipelines \cite{machidon2025}, highlighting the need for more flexible and context-aware refinement strategies. In this setting, VLMs act as intelligent post-processing modules that leverage contextual reasoning and structural constraints to validate OCR outputs.

This paradigm is supported by recent works such as CLOCR-C, which demonstrate that incorporating contextual language models can substantially reduce character error rates and enable reconstruction of degraded text \cite{Bourne2024}. Similarly, \cite{Dai2025} highlight the effectiveness of language-guided correction in structured recognition tasks. Thus, VLMs function as a semantic consistency layer that complements, rather than replaces, visual recognition models in challenging OCR tasks.  

\section{Data Collection}
\label{sec:data-collection}

\begin{figure}[!htbp]
\centering
\includegraphics[width=0.7\textwidth]{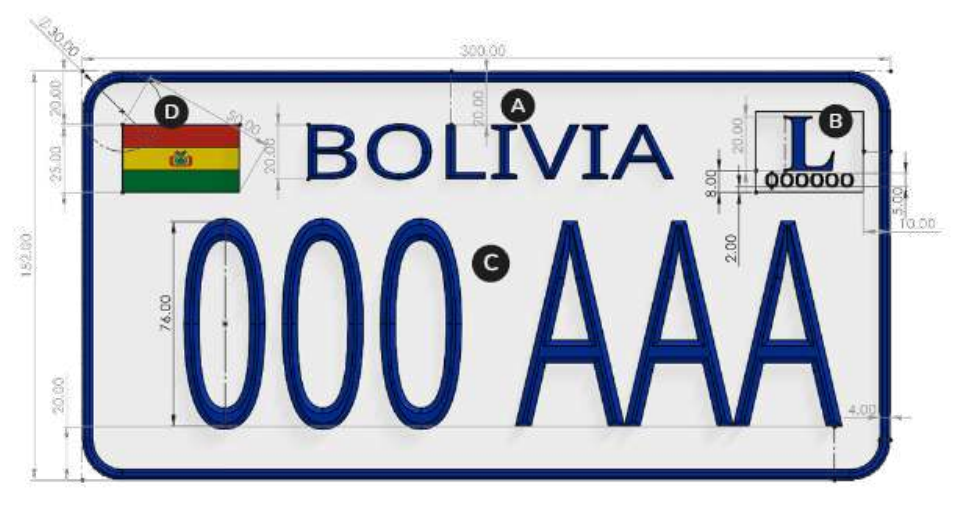}
\caption{Physical details of the Bolivian license plate defined by the Bolivian government (RUAT) \cite{ruat2026}.}
\label{fig:bolivian-license-plate}
\end{figure}

The Bolivian license plate standard, introduced in 1997, consists of three to four digits followed by three letters, rendered in blue on a white background, with the word ``BOLIVIA'' at the top. Vehicles are categorized by department and fuel type, with the department indicated in the upper right corner by a code character. Following 2024 legislation, fuel type determines the plate color scheme: electric or flex-fuel vehicles feature white letters on a blue or green background, respectively. However, these regulations are often not enforced in practice, resulting in most license plates having a white background, as shown in Figure \ref{fig:bolivian-license-plate}.

\begin{figure}[!htbp]
\centering
\makebox[\textwidth][c]{%
\subfloat[]{%
    \includegraphics[height=0.22\textheight,keepaspectratio]{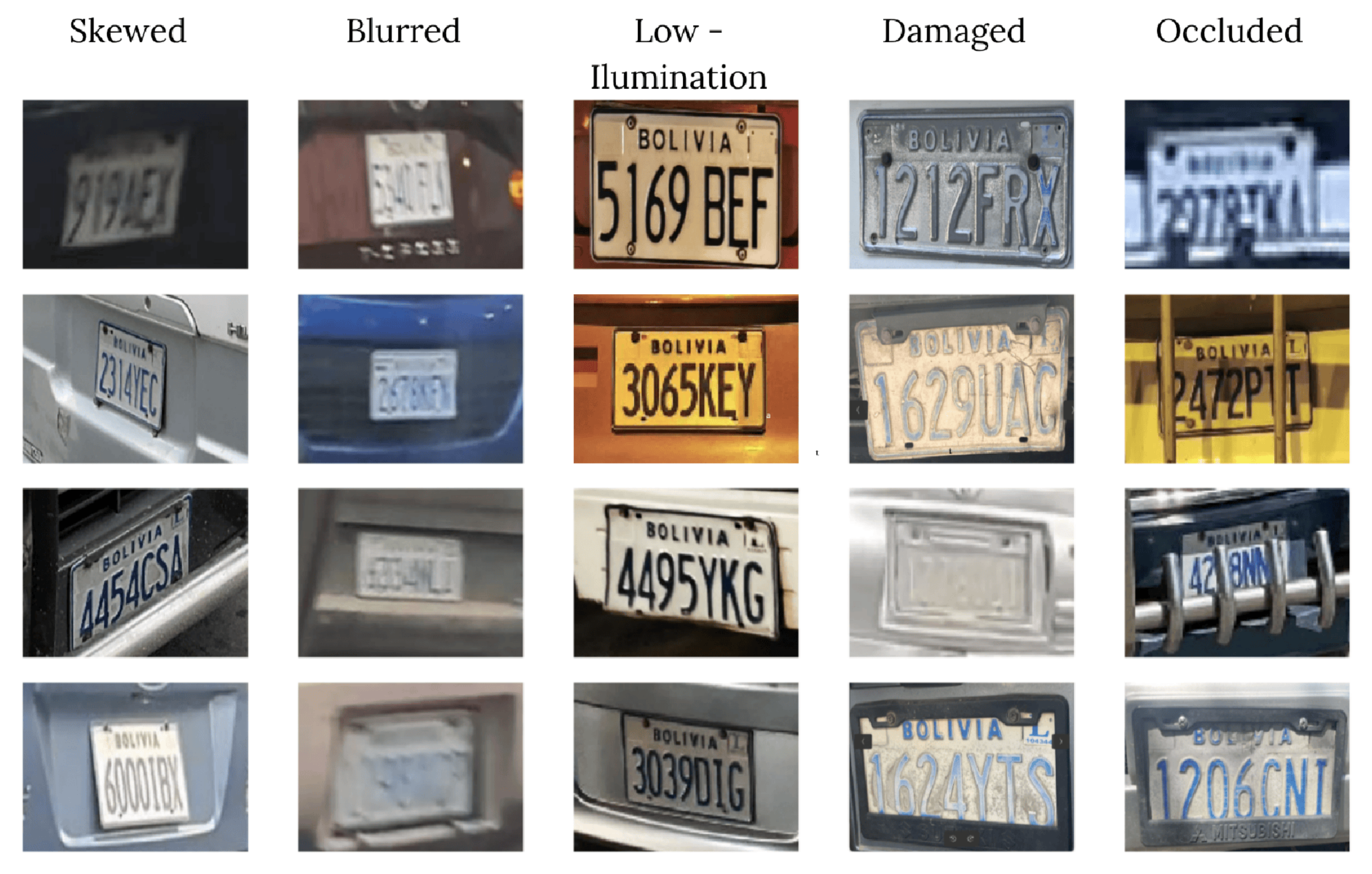}%
    \label{fig:plate-conditions-samples}%
}\hspace{0.05\textwidth}%
\subfloat[]{%
    \includegraphics[height=0.22\textheight,keepaspectratio]{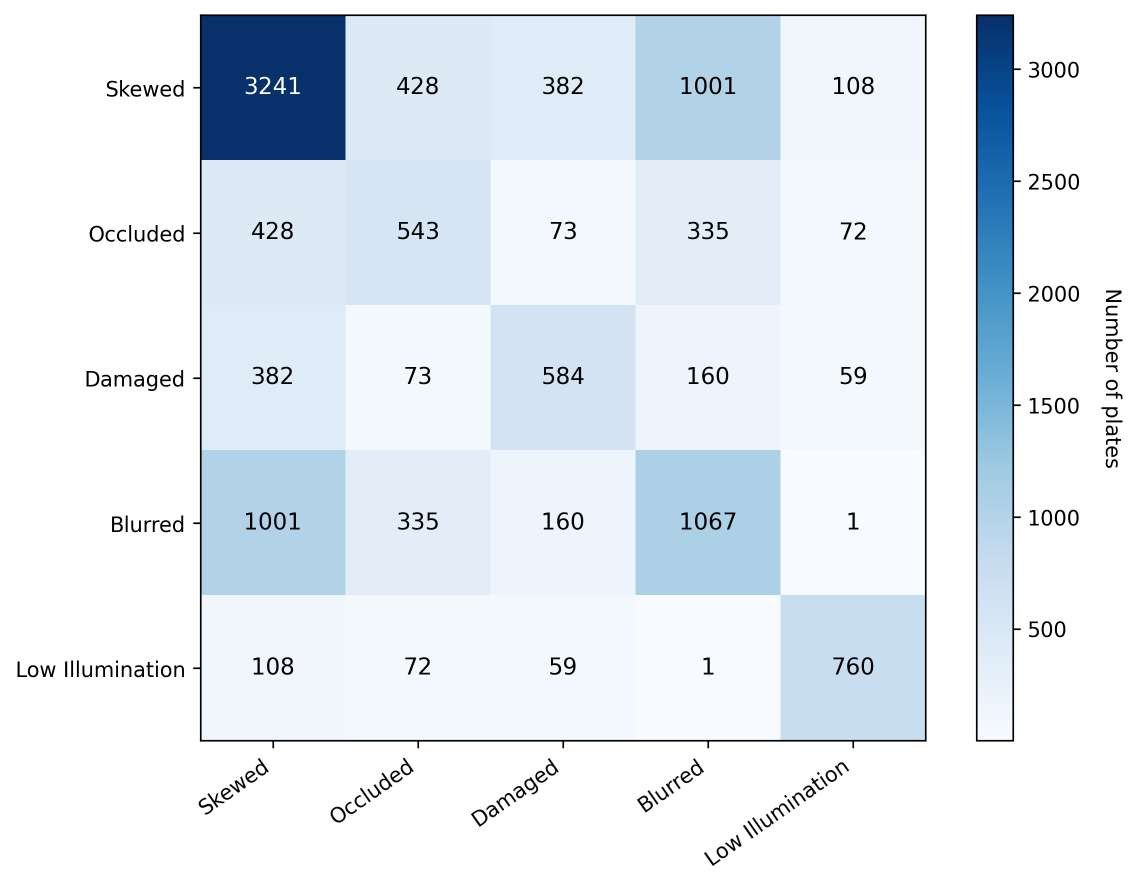}%
    \label{fig:plate-conditions-matrix}%
}%
}

\caption{Analysis of anomalous license plate conditions in the BLPR-A dataset. 
(a) Representative examples of anomalous license plates. 
(b) Co-occurrence matrix showing the frequency of individual and overlapping anomalous conditions.}
\label{fig:anomaly-analysis}
\end{figure}

The implementation of a novel LPDR system for the Bolivian case required a dataset for car and license plate detection under real-world conditions. To this end, we constructed BLPR-A by capturing 3,541 street images in La Paz, Bolivia. Cars, excluding buses and motorcycles, and license plates were annotated using the Roboflow tool. Furthermore, license plate appearances were categorized into normal and anomalous cases (see Figure \ref{fig:plate-conditions-samples}). Figure \ref{fig:plate-conditions-matrix} summarizes the co-occurrence of these anomalies, showing that skewed plates are the most prevalent condition (3,241 instances), followed by blurred plates (1,067), low-illumination cases (760), damaged plates (584), and occluded plates (543). The matrix also shows that skewness frequently co-occurs with blur (1,001 instances), suggesting that viewpoint distortion and image degradation often appear together in real-world captures. In contrast, low illumination occurs mostly as an isolated condition, with limited overlap with blur or other anomalies. Given this distribution, focusing on skewed plates and low-illumination conditions is well justified, as they represent dominant challenges in BLPR-A, whereas blurred cases may require fundamentally different solutions, such as faster shutter speeds.

\begin{figure}[!htbp]
    \centering
    \includegraphics[width=0.8\linewidth]{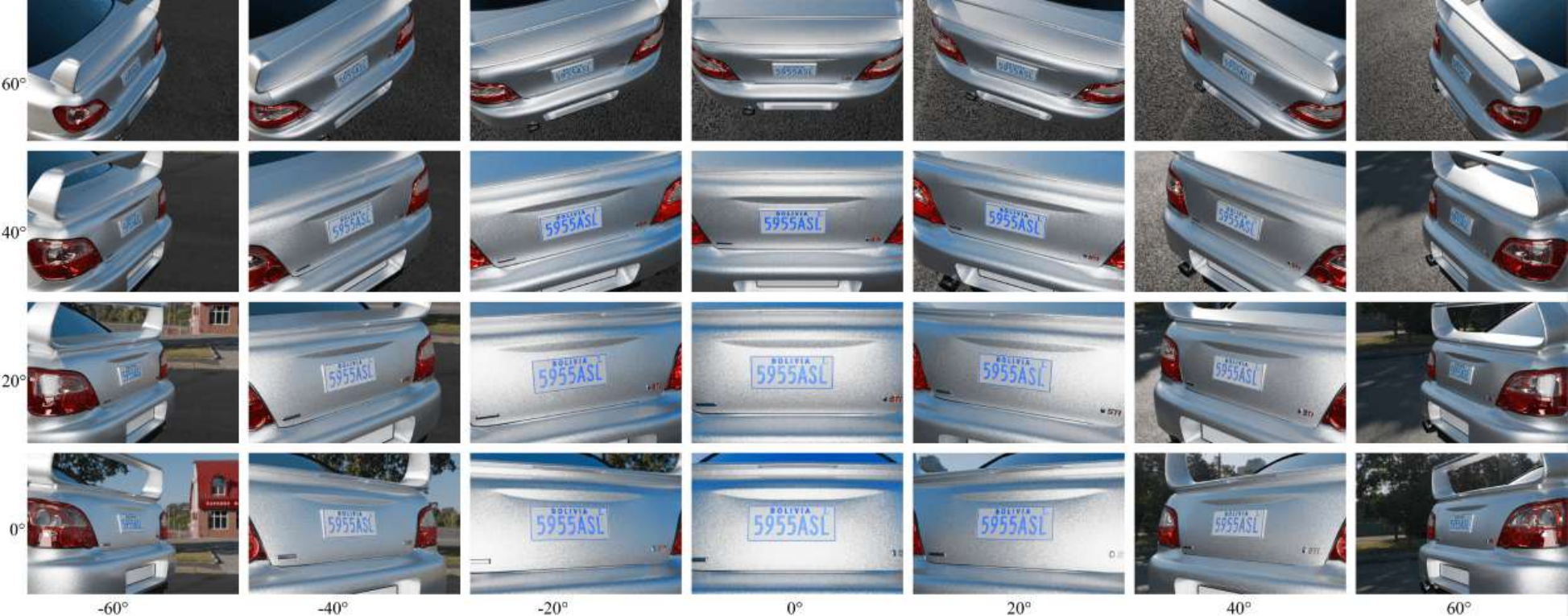}
    \caption{Perspective variations from the BLPR-B dataset, annotated with the corresponding angles (in degrees) of each viewpoint.}
    \label{fig:perspective-variations}
\end{figure}

\begin{figure}[!htbp]
    \centering
    \includegraphics[width=0.5\linewidth]{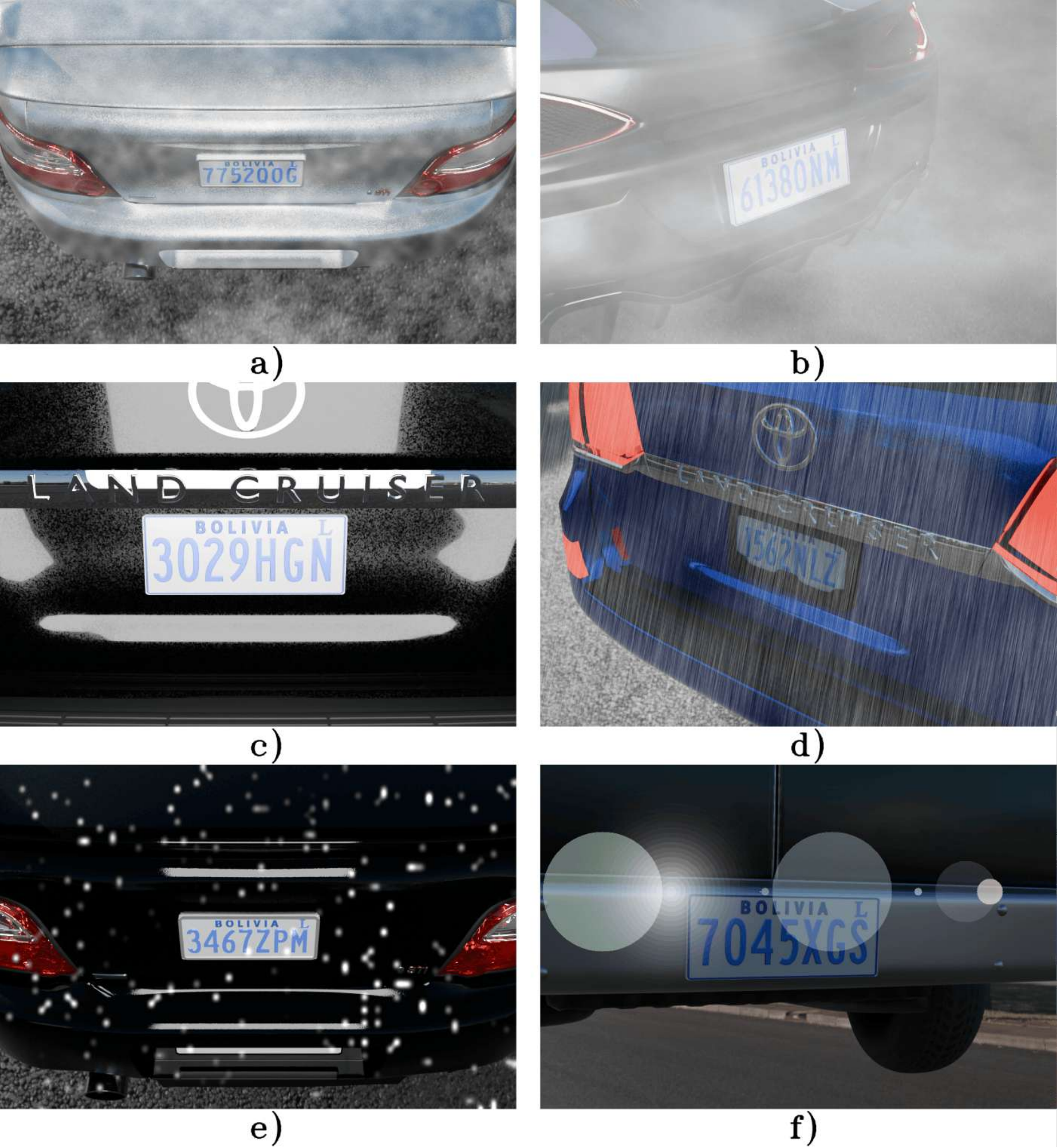}
    \caption{Samples from the BLPR-B dataset with different illumination variations: (a) sun flare, (b) fog, (c) snowflakes, (d) rain, (e) cloudy conditions, and (f) luminosity variation.}
    \label{fig:weather-based-variations}
\end{figure}

Inspecting and annotating BLPR-A revealed a lack of samples with challenging viewpoint and illumination variations. Therefore, we generated a synthetic dataset, namely BLPR-B, using the 3D computer graphics software Blender. The dataset comprises 5,292 license plate images featuring 36 unique plate numbers and five different vehicle models, each rendered in three distinct colors. Perspective variations were systematically introduced, as illustrated in Figure \ref{fig:perspective-variations}, where the horizontal viewing angle ranges from $150^\circ$ to $30^\circ$, and the vertical angle ranges from $150^\circ$ to $90^\circ$. In addition, illumination variations were generated by introducing a light source in Blender at a fixed distance from the camera and varying its intensity. Moreover, an obstructed subset containing 1,764 images was generated. To expand the synthetic dataset, six weather conditions were simulated: rain, snow, cloudy conditions, fog, sun glare, and snowfall (see Figure \ref{fig:weather-based-variations}). For each environmental condition, 1,764 images were produced from the obstructed subset and 5,292 images from the original unobstructed dataset, yielding a total of 42,336 augmented samples. Overall, the final BLPR-B dataset comprises 49,392 images.

\begin{figure}[!htbp]
    \centering
    \includegraphics[width=0.5\linewidth]{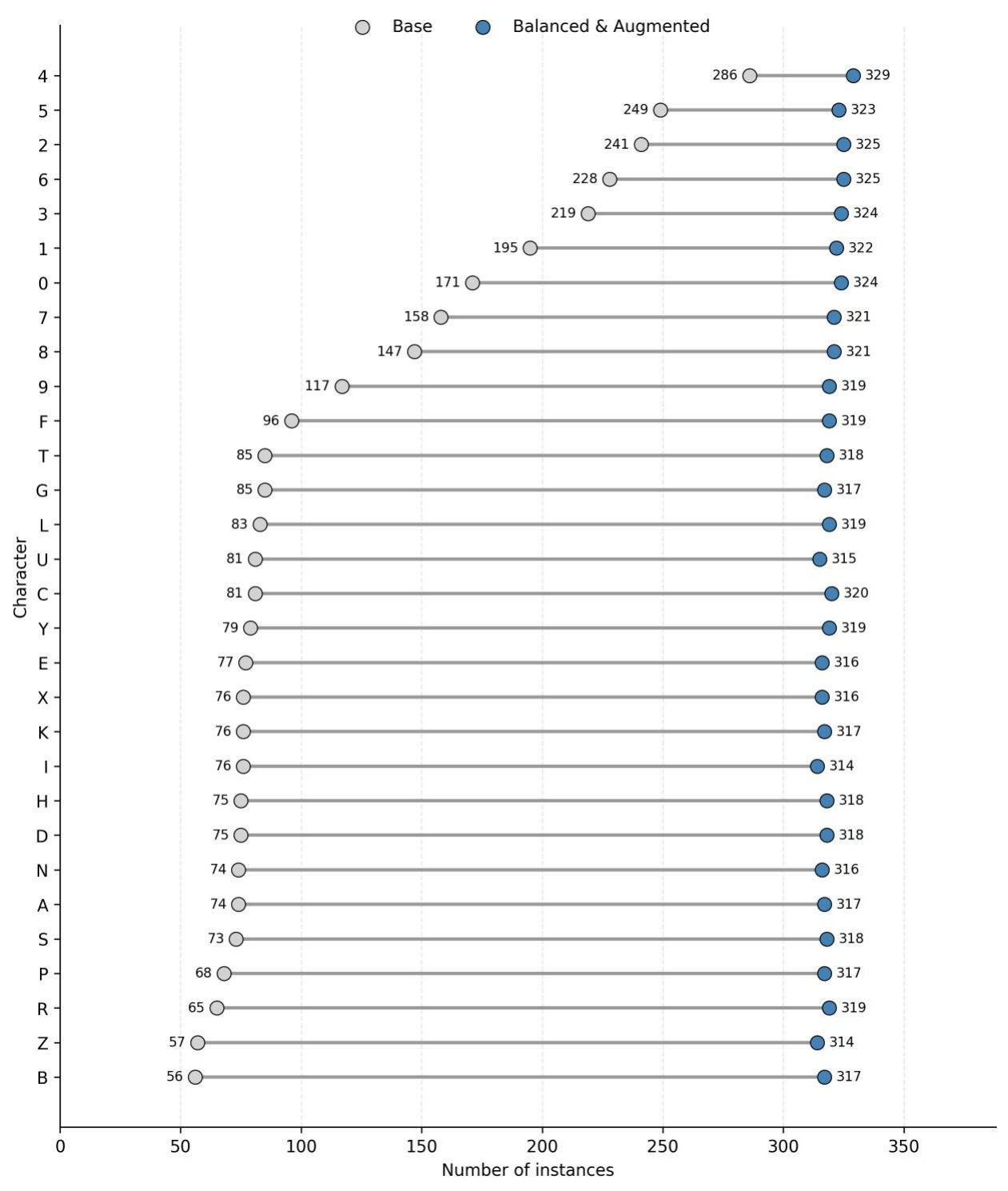}
    \caption{Per-character distribution in BLPR-C before and after balancing.}
    \label{fig:blpr-c-composition}
\end{figure}

Subsequently, we collected 520 images in La Paz, Bolivia, for the OCR module of the framework, each containing a unique license plate, and annotated the individual characters for object detection. This dataset, designated as BLPR-C, exhibited a class imbalance ratio of 5.11 and a normalized entropy of 0.959 (see Figure \ref{fig:blpr-c-composition}). To address this, a balancing-focused data augmentation strategy was developed, in which ROIs from underrepresented classes were extracted and patched into the bounding boxes of more frequent categories. This augmentation phase reduced the imbalance ratio to 1.05 and increased the normalized entropy to 0.998, resulting in a total of 1,374 images. To further simulate challenging scenarios, a secondary augmentation process was applied to the balanced dataset, in which each plate was expanded into six variants, including clockwise and counterclockwise tilting and shearing, Gaussian blurring, and the addition of Gaussian noise. This process resulted in a total of 9,615 annotated license plate images.

Finally, an additional dataset, BLPR-D, was created for independent validation under controlled acquisition conditions. This dataset consists of street-level images captured from different perspectives and under varying lighting conditions. Each image is accompanied by a label that includes the image name, ambient luminance measured in lux, plate annotations in string format, and metadata describing the plate's distance and angle relative to the camera. The number of original samples and characteristics of the four datasets (BLPR-A, BLPR-B, BLPR-C, and BLPR-D) are shown in Table \ref{tab:lpr-regional-datasets}.

\section{Proposed BLPR System}
\label{sec:proposed-system}

\begin{figure}[!htbp]
    \centering\includegraphics[width=\linewidth]{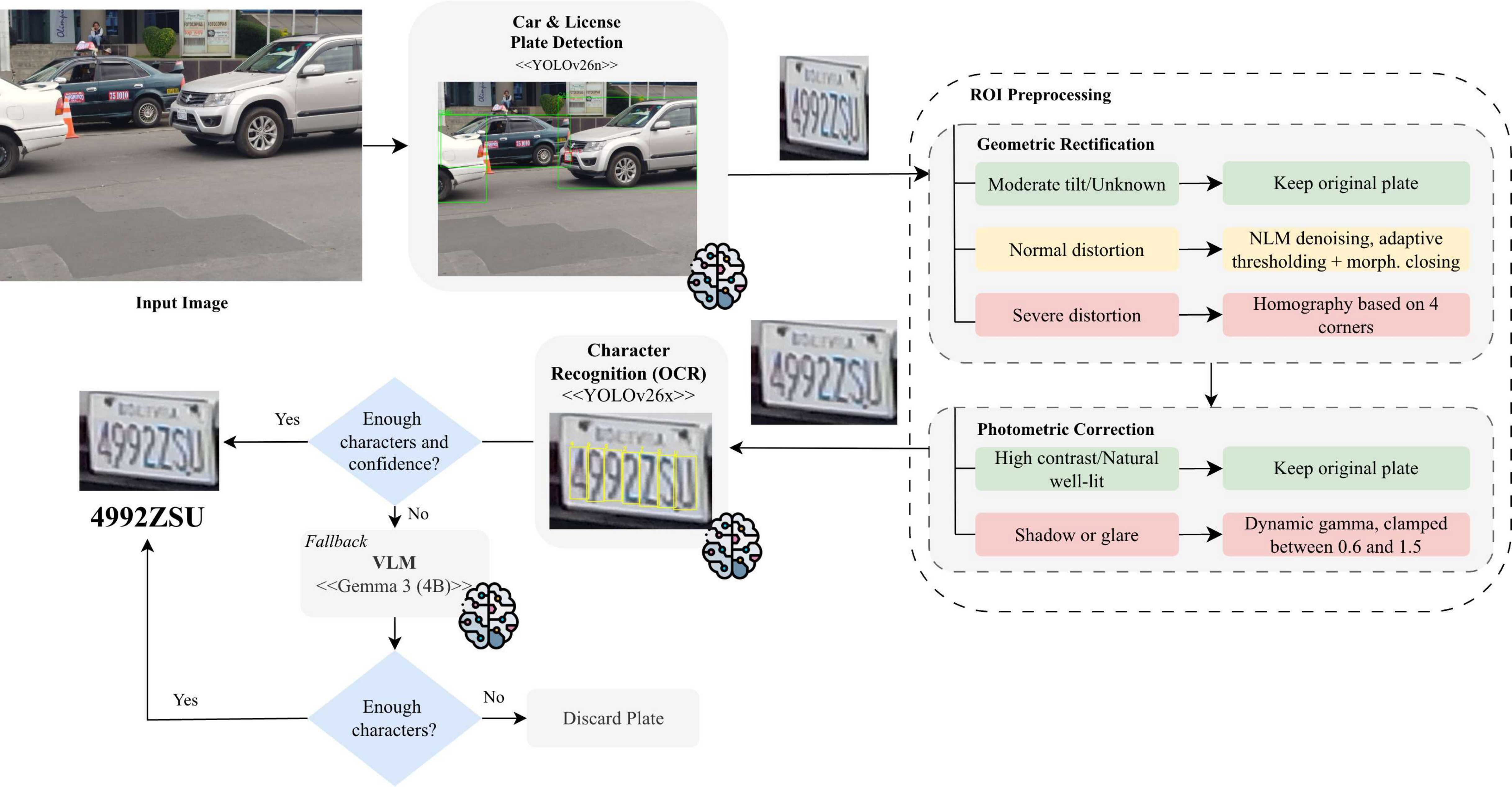}
    \caption{BLPR system proposed for license plate recognition under varying viewpoints and illumination conditions.}
    \label{fig:pipeline}
\end{figure}

As shown in Figure \ref{fig:pipeline}, the proposed system comprises three main modules: (1) a car and license plate detector, (2) a preprocessing pipeline for skew and illumination correction, and (3) an OCR module. The overall objective of the system is to map an input image to one or more sequences of alphanumeric characters corresponding to detected license plates. Formally, let $I \in \mathbb{R}^{H \times W \times 3}$ denote an input image. The detection module produces a set of ROIs, each corresponding to a license plate. Each ROI is processed independently, and the goal is to predict a sequence of characters $y = (c_1, \dots, c_n)$ for a given ROI. The proposed pipeline for a single ROI can be expressed as:

\begin{equation}
y = f_{\text{OCR}} \left( f_{\text{photo}} \left( f_{\text{rect}} (R) \right) \right)
\end{equation}

where $R \in \mathbb{R}^{h \times w \times 3}$ denotes a region of interest, $f_{\text{rect}}$ represents geometric rectification, $f_{\text{photo}}$ denotes photometric correction, and $f_{\text{OCR}}$ corresponds to the character recognition module. The following sections describe each component of the BLPR system, while a detailed pseudo-code of the system is provided in Algorithm \ref{alg:alpr_pipeline_simplified}.

\begin{algorithm}[!htbp]
\small
\caption{BLPR system}\label{alg:alpr_pipeline_simplified}
\begin{algorithmic}[1]
\Require Input frame: $I$
\Procedure{LPR\_System}{$I$}
    
    \Statex \textbf{Module 1: Plate Detection (LPD)}
    \State $Cars, Plates \gets \text{YOLO\_LPD}(I, \text{conf} \ge 0.50)$
    \State $ValidPlates \gets \{p \in Plates \mid \exists c \in Cars : \text{Inside}(p, c)\}$
    \If{$ValidPlates$ is empty} \Return \textbf{null} \EndIf
    \State $ROI_{raw} \gets \text{Crop}(I, \text{GetBestBox}(ValidPlates) + 10px)$

    \Statex
    \Statex \textbf{Module 2: Pre-processing Pipeline}
    \State $ROI_{leveled} \gets \Call{GeometricRectification}{ROI_{raw}}$
    \State $ROI_{opt} \gets \Call{PhotometricCorrection}{ROI_{leveled}}$

    \Statex
    \Statex \textbf{Module 3: OCR}
    \State $Chars \gets \text{YOLO\_LPR}(ROI_{opt})$
    \State $Chars \gets \text{FilterIgnoredClasses}(Chars, \{\text{``BOLIVIA''}, \text{``\_''}\})$
    \State $final\_text, min\_conf \gets \text{ExtractSortedText}(Chars)$
    \State $length \gets \text{Count}(Chars)$

    \Statex
    \Statex \textbf{Module 4: VLM Fallback}
    \If{$length < 6 \textbf{ or } (length > 0 \textbf{ and } min\_conf < 0.15 \cdot \max(Chars.conf))$}
        \State $Prompt \gets \text{``Read plate. Alphanumeric only. No BOLIVIA.''}$
        \State $final\_text \gets \Call{Gemma3\_4B}{ROI_{opt}, Prompt}$
    \EndIf

    \State \Return $final\_text$
\EndProcedure
\end{algorithmic}
\end{algorithm}

\textbf{Car and License Plate Detection:}
The pipeline begins with a car and license plate detection module based on a YOLO26n network, which localizes and extracts license plate regions from the input image. To improve robustness under diverse environmental conditions, the detector is first pretrained on the synthetic BLPR-B dataset and subsequently fine-tuned on the real-world BLPR-A dataset. This two-stage training strategy enhances generalization to viewpoint distortions and illumination variability. During inference, a confidence threshold of 0.50 is applied to filter detections; this value was empirically selected to balance recall and false positives in cluttered scenes while preserving real-time constraints.

\begin{table}[!htbp]
\centering
\caption{Sensitivity analysis of geometric rectification parameters.}
\label{tab:rect_var_comp}
\begin{adjustbox}{max width=\textwidth}
\begin{tabular}{lcccccccc}
\toprule
\textbf{Modification}                            & \textbf{Value}                    & \textbf{Average Similarity}         & \textbf{Precision}                  & \textbf{Recall}                     & \textbf{F1 Score (Normal)}          & \textbf{F1 Score (Tilted)}          & \textbf{F1 Score (Steep)}           & \textbf{Plates Rectified}           \\ 
\midrule
                                                 & \cellcolor{gray!12}\textbf{1.05}  & \cellcolor{gray!12}\textbf{0.8979}  & \cellcolor{gray!12}\textbf{0.9109}  & \cellcolor{gray!12}\textbf{0.8965}  & \cellcolor{gray!12}\textbf{0.8909}  & \cellcolor{gray!12}\textbf{0.9423}  & \cellcolor{gray!12}\textbf{0.8848}  & \cellcolor{gray!12}\textbf{9.6\%}   \\ 
                                                 & 1.10                              & 0.8951                              & 0.9089                              & 0.8934                              & 0.8900                              & 0.9395                              & 0.8808                              & 8.7\%                               \\
                                                 & 1.15                              & 0.8928                              & 0.9082                              & 0.8910                              & 0.8853                              & 0.9358                              & 0.8842                              & 5.1\%                               \\
                                                 & 1.20                              & 0.8933                              & 0.9088                              & 0.8913                              & 0.8831                              & 0.9385                              & 0.8855                              & 4.5\%                               \\
\multirow{-5}{*}{Foreshortening Threshold $r$}   & 1.25                              & 0.8949                              & 0.9089                              & 0.8931                              & 0.8895                              & 0.9365                              & 0.8833                              & 4.3\%                               \\ 
\midrule
                                                 & \cellcolor{gray!12}\textbf{10º}   & \cellcolor{gray!12}\textbf{0.8981}           & \cellcolor{gray!12}\textbf{0.9129}  & \cellcolor{gray!12}0.8958           & \cellcolor{gray!12}0.8882           & \cellcolor{gray!12}\textbf{0.9396}  & \cellcolor{gray!12}\textbf{0.8918}  & \cellcolor{gray!12}\textbf{5.3\%}   \\ 
                                                 & 12.5º                             & 0.8955                              & 0.9109                              & 0.8931                              & 0.8914                              & 0.9353                              & 0.8852                              & 5.1\%                               \\
                                                 & 15º                               & 0.8943                              & 0.9083                              & 0.8925                              & 0.8886                              & 0.9370                              & 0.8821                              & 5.1\%                               \\
                                                 & 17.5º                             & \textbf{0.8981}                              & 0.9115                              & \textbf{0.8962}                     & \textbf{0.8924}                     & 0.9375                              & 0.8876                              & 5.1\%                               \\
\multirow{-5}{*}{Tilt-Angle Threshold $\theta$}  & 20º                               & 0.8935                              & 0.9076                              & 0.8916                              & 0.8831                              & 0.9349                              & 0.8874                              & 5.1\%                               \\ 
\midrule
                                                 & 15\%                              & 0.8964                              & 0.9108                              & 0.8946                              & 0.8878                              & 0.9353                              & 0.8912                              & 4.9\%                               \\
                                                 & \cellcolor{gray!12}\textbf{20\%}  & \cellcolor{gray!12}\textbf{0.9003}           & \cellcolor{gray!12}\textbf{0.9139}  & \cellcolor{gray!12}\textbf{0.8983}  & \cellcolor{gray!12}0.8886           & \cellcolor{gray!12}0.9376           & \cellcolor{gray!12}\textbf{0.8982}  & \cellcolor{gray!12}4.9\%            \\
                                                 & 25\%                              & 0.8936                              & 0.9080                              & 0.8919                              & 0.8843                              & 0.9360                              & 0.8860                              & 5.1\%                               \\
                                                 & 30\%                              & 0.8938                              & 0.9078                              & 0.8928                              & 0.8828                              & \textbf{0.9408}                     & 0.8847                              & 5.1\%                               \\
\multirow{-5}{*}{Width Guardrail}                & 35\%                              & 0.8972                              & 0.9106                              & 0.8958                              & \textbf{0.8891}                     & 0.9343                              & 0.8921                              & \textbf{6.0\%}                      \\ 
\bottomrule
\end{tabular}
\end{adjustbox}
\end{table}

\begin{table}[!htbp]
\centering
\caption{Performance metrics for the illumination sub-pipeline under varying configurations.}
\label{tab:illumination_var_comp}
\begin{adjustbox}{max width=\textwidth}
\begin{tabular}{llccccc}
\toprule
\textbf{Modification}                        & \textbf{Value}                               & \textbf{Average Similarity}        & \textbf{Precision}                 & \textbf{Recall}                    & \textbf{F1 Score}                  & \textbf{Illuminated Images}         \\
\midrule
\multirow{4}{*}{Contrast Threshold $\sigma$} & 40                                           & 0.8953                             & 0.9081                             & 0.8937                             & 0.9009                             & 5.3\%                               \\ 
                                             & \cellcolor{gray!12}\textbf{50}               & \cellcolor{gray!12}\textbf{0.8958} & \cellcolor{gray!12}0.9088          & \cellcolor{gray!12}\textbf{0.8949} & \cellcolor{gray!12}\textbf{0.9018} & \cellcolor{gray!12}18.1\%           \\ 
                                             & 60                                           & 0.8944                             & 0.9080                             & 0.8928                             & 0.9004                             & 27.7\%                              \\
                                             & 70                                           & 0.8942                             & \textbf{0.9094}                    & 0.8922                             & 0.9007                             & \textbf{31.9\%}                     \\
\midrule
\multirow{3}{*}{Brightness Range $\mu$}      & min=70, max=150                              & 0.8887                             & 0.9030                             & 0.8870                             & 0.8949                             & \textbf{34.7\%}                     \\
                                             & min=80, max=160                              & \textbf{0.8944}                    & 0.9078                             & \textbf{0.8931}                    & 0.9004                             & 27.2\%                              \\ 
                                             & \cellcolor{gray!12}\textbf{min=90, max=170}  & \cellcolor{gray!12}\textbf{0.8944} & \cellcolor{gray!12}\textbf{0.9095} & \cellcolor{gray!12}0.8928          & \cellcolor{gray!12}\textbf{0.9010} & \cellcolor{gray!12}17.4\%           \\
\midrule
\multirow{3}{*}{Gamma Clamp $\gamma$}        & min=0.7, max=1.4                             & 0.8951                             & \textbf{0.9097}                             & 0.8928                             & 0.9012                             & 27.4\%                              \\ 
                                             & \cellcolor{gray!12}\textbf{min=0.6, max=1.5} & \cellcolor{gray!12}\textbf{0.8952} & \cellcolor{gray!12}0.9095 & \cellcolor{gray!12}\textbf{0.8937} & \cellcolor{gray!12}\textbf{0.9016} & \cellcolor{gray!12}\textbf{27.7\%}  \\ 
                                             & min=0.5, max=1.6                             & 0.8942                             & 0.9094                             & 0.8919                             & 0.9005                             & \textbf{27.7\%}                     \\  
\bottomrule
\end{tabular}
\end{adjustbox}
\end{table}

\textbf{Geometric Rectification:} Following detection, the ROI is passed to a Geometric Rectification module, which acts as a router with three processing branches. Its purpose is to correct 3D foreshortening and 2D tilt while avoiding interpolation artifacts. By default, the module employs a ``Scout'' process that generates a high-contrast version of the ROI using Contrast Limited Adaptive Histogram Equalization (CLAHE), thereby assisting the Canny edge detector under low-light conditions. The largest contour is identified, approximated as a four-point polygon, and used to compute the foreshortening ratio $r$ and the top-edge tilt angle $\theta$. These are defined as (i) the ratio between the lengths of the top and bottom edges of the quadrilateral and (ii) the angle between the top edge and the horizontal axis, respectively.

As shown in Table \ref{tab:rect_var_comp}, we empirically analyzed the sensitivity of the pipeline to three main rectification parameters used in our algorithm: the Foreshortening Ratio Threshold $r$, the Tilt-Angle Threshold $\theta$, and the Width Guardrail, under different viewing geometries. The Foreshortening Ratio Threshold exhibited a clear inverse relationship with rectification quality; a lower threshold of 1.05 produced the highest average similarity score (0.8979) and a peak rectification rate of 9.6\%, indicating that stricter constraints on foreshortening lead to more accurate geometric corrections. Performance with respect to the Tilt-Angle Threshold remained stable across the tested range, although precision was highest at a threshold of 10$^{\circ}$ (0.9129), while recall and F1 score were highest at 17.5$^{\circ}$ for ‘Normal’ plate orientations. Finally, the Width Guardrail analysis showed that a setting of 20\% achieved the best performance on the core accuracy metrics, with a peak average similarity of 0.9003.

Subsequent processing depends on three scenarios. If $r > 1.05$ or $\theta > 10.0^\circ$, a homography transformation is applied to rectify the plate, provided the transformation does not exceed a 20\% width guardrail designed to prevent geometric artifacts. For plates with low distortion, Non-Local Means (NLM) denoising and morphological closing are applied to refine the text region based on its solidity and area, whereas moderate cases are left unmodified to avoid unnecessary interpolation and preserve natural image gradients. These values were selected empirically to balance distortion correction and artifact avoidance. The algorithm and corresponding parameter definitions are detailed in Appendix~\ref{anx:preprocessing-algorithms}.

\textbf{Illumination Correction:} The enhanced ROI then undergoes photometric correction to address lighting artifacts such as severe shadows or glare. This stage, detailed in Appendix \ref{anx:illumination-correction}, converts the image to the HSV color space and analyzes the Value channel to estimate the mean brightness $\mu$ and contrast $\sigma$. If the image already exhibits sufficient contrast ($\sigma > 50$) or maintains a balanced brightness level ($90 \leq \mu \leq 170$), correction is skipped to preserve the original image quality. Otherwise, for images affected by shadow or glare, dynamic gamma correction is applied. Specifically, a gamma multiplier $\gamma$ is computed and clamped within the range $0.6 \leq \gamma \leq 1.5$ to adjust brightness toward a neutral level while preserving the original Hue and Saturation channels.

These thresholds were selected to provide stable enhancement and robustness while avoiding overcorrection, as shown in Table \ref{tab:illumination_var_comp}. Among the evaluated parameters, the Contrast Threshold had the strongest effect on processing throughput. A threshold of 50 achieved the highest F1 score (0.9018), whereas increasing this value to 60 and 70 substantially raised the proportion of illuminated images to 27.7\% and 31.9\%, respectively, with only a marginal reduction in peak accuracy. The brightness range analysis revealed a clear trade-off between coverage and quality. Although the range [70, 150] maximized the illuminated volume (34.7\%), it produced the lowest average similarity (0.8887). In contrast, the range [90, 170] achieved the highest precision (0.9095) and average similarity (0.8944), indicating better character preservation under low-light conditions. Finally, the Gamma Clamp performed best with a configuration of [0.6, 1.5], achieving the highest similarity (0.8952) and F1 score (0.9016) while maintaining high image throughput.

\textbf{Optical Character Recognition:} Character extraction is then performed using a YOLO26x model trained on the BLPR-C dataset. The model resizes the input images to a resolution of $1024 \times 1024$ and outputs bounding boxes and confidence scores for each detected character. During development, we identified the importance of detecting the word ``Bolivia'' on the plates; therefore, an additional class corresponding to this text was included in the BLPR-C dataset. After inference, this class is filtered out, and the remaining characters are grouped into lines using a 50\% vertical overlap criterion and sorted from left to right within each line to produce the final text string, along with the total character count and associated confidence scores. This process also removes the upper-right character corresponding to the department code (e.g., L for La Paz).

To improve robustness in challenging scenarios without replacing the primary OCR pathway, a conditional VLM fallback mechanism is employed. A ``confidence tripwire'' activates this fallback when the YOLO26x model detects fewer than six characters or when a character's relative confidence falls below a threshold with respect to the maximum confidence within the plate. Formally, the fallback is triggered if
\begin{equation}
\frac{\min_i c_i}{\max_i c_i} < \tau,
\end{equation}
where $c_i$ denotes the confidence of the $i$-th character and $\tau=0.15$ is a predefined threshold. When triggered, the preprocessed ROI is encoded as a JPEG image and passed to a Gemma3 model with 4 billion parameters (4B), which is prompted to return only the alphanumeric characters present on the plate while excluding metadata, formatting, and country identifiers.

The threshold $\tau$ directly affects the trade-off between OCR accuracy and system latency. Table \ref{tab:confidence_comparison} shows that by raising the threshold from 0.10 to 0.20, the peak Average Levenshtein Similarity achieved was 0.8956 at a 456 ms inference time, demonstrating that a more stringent activation of the fallback mechanism successfully discards predictions with low confidence in order to prioritize stronger secondary processing. The lowest latency (392 ms) was observed at $\tau = 0.15$, which also tied for the lowest activation rate. Furthermore, the difference in similarity of 0.006 was considered negligible. Beyond a threshold of 0.20, the Fallback Activation Rate continued increasing to 0.379, yet similarity scores started decreasing, indicating a point of diminishing returns where the fallback mechanism is unnecessarily activated for higher-quality primary detections. Consequently, a Confidence Threshold of 0.15 was selected as the optimal operating point for the final pipeline.

\begin{table}[!htbp]
\centering
\caption{Comparative analysis of OCR performance metrics across varying Confidence Thresholds.}
\label{tab:confidence_comparison}
\begin{tabular}{lccc}
\toprule
\textbf{\begin{tabular}[c]{@{}l@{}}Confidence\\ Threshold $\tau$\end{tabular}} & \textbf{\begin{tabular}[c]{@{}l@{}}Average Levenshtein\\ Similarity\end{tabular}} & \textbf{\begin{tabular}[c]{@{}l@{}}Fallback Activation\\ Rate\end{tabular}} & \textbf{Latency (ms)} \\ 
\midrule
0.10          & 0.8888          & \textbf{0.294} & 399  \\
0.20          & \textbf{0.8956} & 0.349          & 456  \\ 
0.25          & 0.8933          & 0.379          & 494  \\
\midrule
\rowcolor{gray!12}
\textbf{0.15} & 0.8897          & \textbf{0.294} & \textbf{392}\\
\bottomrule
\end{tabular}
\end{table}

\begin{table}[!htbp]
\centering
\caption{Comparison of detection models on the BLPR-A test set at a confidence threshold of 0.5 and IoU of 0.5.}
\label{tab:detector-comparison}
\begin{tabular}{lcccccc}
\hline
\textbf{Model} & \textbf{Precision} & \textbf{Recall} & \textbf{F1 Score} & \textbf{mAP@0.5} & \textbf{mAP@0.5:0.95} & \textbf{GFLOPs}\\
\hline
YOLOv8n & 0.942 & 0.875 & 0.905 & 0.924 & 0.757 & 8.7 \\
YOLOv8s & \textbf{0.943} & \textbf{0.924} & \textbf{0.933} & 0.948 & \textbf{0.794} & 21.5 \\
YOLOv11s & \textbf{0.943} & 0.900 & 0.920 & 0.942 & 0.792 & 21.5 \\

\midrule
\rowcolor{gray!12}
\textbf{YOLO26n} & 0.923 & 0.889 & 0.906 & \textbf{0.979} & 0.782 & \textbf{5.4} \\
\bottomrule
\hline
\end{tabular}
\end{table}

Unlike conventional LPDR pipelines, the proposed system introduces a dynamic rectification routing strategy and a confidence-triggered fallback to improve robustness while preserving efficiency. The method may fail in scenarios involving severe motion blur, low-fidelity imagery, or when the license plate is not detected during the initial ROI extraction.

\section{Experimental Results}
\label{sec:experimental-results}

\subsection{Vehicle, License Plate, and Character-Level Detection}

The experimental setup of the proposed BLPR system consisted of a workstation equipped with an AMD Ryzen 9 7950X3D processor, 32 GB of RAM, and an NVIDIA GeForce RTX 4080 SUPER GPU with 16 GB of VRAM. The car and license plate detection models were trained for 100 epochs with an input image size of 640 × 640 pixels and a confidence threshold of 0.50. Table \ref{tab:detector-comparison} presents the results of all the evaluated YOLO architectures, demonstrating that all of them reach performance metrics that satisfy the operational requirements for the detection task, with each model achieving precision above 0.92 and mAP@0.5 above 0.92. However, for this implementation, YOLO26n was chosen as the best backbone due to its better computational efficiency.

Although models such as YOLOv8s and YOLOv11s offer high F1 scores, YOLO26n attains the highest mAP@0.5 (0.979) with a substantial reduction in computational demand. In particular, YOLO26n only requires 5.4 GFLOPs, which is a significant reduction in complexity compared to the 21.5 GFLOPs required by YOLOv8s and YOLOv11s. This reduction in GFLOPs is necessary to enable real-time processing and feasibility of deployment in resource-constrained settings while still maintaining localization accuracy. Furthermore, during experimentation, we observed the importance of adding a 10-pixel padding to the predicted license plate bounding box to prevent character truncation during ROI extraction. This configuration resulted in an mAP@50–95 score of 0.7818. Moreover, the model achieved recall and precision of 0.8890 and 0.9227, respectively, and an F1-score of 0.9055, indicating robust and balanced performance in object localization across multiple IoU thresholds.

\begin{table}[!htbp]
\small
\centering
\caption{Comparison of OCR models on the BLPR-C (controlled) and BLPR-D (real-world) datasets. YOLO26x achieves the best precision–recall trade-off and overall performance, particularly under real-world conditions.}
\label{tab:ocr-comparison}
\begin{adjustbox}{max width=\textwidth}
\begin{tabular}{lcccccccccc}
\hline
\multirow{2}{*}{\textbf{Model}} & \multicolumn{2}{c}{\textbf{Average Similarity}} & \multicolumn{2}{c}{\textbf{Precision}} & \multicolumn{2}{c}{\textbf{Recall}} & \multicolumn{2}{c}{\textbf{F1 Score}} & \multicolumn{2}{c}{\textbf{Time (ms)}} \\
& BLPR-C & BLPR-D & BLPR-C & BLPR-D & BLPR-C & BLPR-D & BLPR-C & BLPR-D & BLPR-C & BLPR-D \\
\hline
EasyOCR     & 0.4035 & 0.5586 & 0.4907 & 0.5379 & 0.3380 & 0.6268 & 0.4003 & 0.5789 & 64 & 31\\
PaddleOCR   & 0.1942 & 0.5218 & 0.2480 & 0.5136 & 0.2178 & 0.6359 & 0.2319 & 0.5682 & 144 & 91\\
TrOCR       & 0.9040 & \textbf{0.7829} & 0.9147 & 0.8158 & 0.8957 & \textbf{0.7764} & 0.9051 & 0.7956 & 61 & 55 \\
YOLO26n     & 0.7401 & 0.4889 & 0.7856 & 0.7107 & 0.7651 & 0.4908 & 0.7752 & 0.5807 & \textbf{6} & \textbf{6}\\
\midrule
\rowcolor{gray!12}
\textbf{YOLO26x} & \textbf{0.9902} & 0.7266 & \textbf{0.9862} & \textbf{0.9128} & \textbf{0.9948} & 0.7132 & \textbf{0.9904} & \textbf{0.8008} & 18 & 18\\
\hline
\end{tabular}
\end{adjustbox}
\end{table}

As shown in Table \ref{tab:ocr-comparison}, we evaluated EasyOCR, PaddleOCR, TrOCR, YOLO26n, and YOLO26x on the BLPR-C and BLPR-D datasets to define the OCR module of the proposed framework. The splits for both datasets are 80\% for training, 10\% for validation and 10\% for testing. The results clearly indicate that YOLO26x provides the best overall performance, particularly in terms of precision–recall balance and robustness across datasets. Notably, YOLO26x achieves near-saturated performance on BLPR-C (F1-score of 0.9904), while maintaining competitive generalization on the more challenging BLPR-D dataset (F1-score of 0.8008), outperforming all competing methods in precision (0.9128). This suggests that the model effectively captures character-level features while remaining resilient to real-world variability. 

The optimal configuration with YOLO26x was obtained by training for 50 epochs with a batch size of 5 and an input resolution of $1024 \times 1024$ pixels. The training process used an initial learning rate ($lr_0$) of 0.001 and a final learning rate factor ($lrf$) of 0.0005. Importantly, data augmentation was deliberately constrained to preserve the class balance introduced in BLPR-C. In particular, mixup was disabled and mosaic augmentation was reduced, as these transformations were observed to introduce unrealistic character compositions that degrade OCR performance. 

Under this configuration, YOLO26x achieved an mAP@50–95 of 0.9848 and an F1-score of 0.9964, with recall (0.9953) and precision (0.9975) indicating a highly reliable detection regime. These results correspond to a very low false-positive rate of 0.25\% and a detection rate of 99.5\% on the test set, demonstrating that the proposed OCR module is both highly accurate and stable under controlled conditions.

\subsection{Vision–Language Models as a Fallback Mechanism}

Preliminary experimental results showed that the Llama 3.2 Vision model, a recent VLM, outperformed approaches that are based solely on character-level object detection (e.g., YOLO, EfficientNet). Therefore, we identified an opportunity to combine both methodologies to enhance the OCR module's accuracy. While the Llama 3.2 Vision model achieves strong performance in image-based text recognition, it incurs high computational costs and often requires specialized hardware for deployment. Consequently, we employ VLMs as a fallback mechanism rather than as the primary OCR model.

To determine the optimal VLM as a fallback mechanism, we experimented with Llama 3.2 Vision, Ministral 3, Gemma3, and DeepSeek-OCR using the BLPR-D dataset (see Table \ref{Model_Val_perf}). These models were evaluated using Average Similarity (Levenshtein distance), precision, recall, F1-score, and model size (in billions of parameters). Specifically, we computed the Levenshtein distance $lev$ to measure the similarity between two strings by determining the minimum number of character edits required to transform one string into the other. The metric is defined as follows:

\begin{equation}
\text{lev}_{a,b}(i,j) =
\begin{cases}
\max(i,j) & \text{if } \min(i,j) = 0 \\
\min \left\{
\begin{array}{l}
\text{lev}_{a,b}(i-1,j) + 1 \\
\text{lev}_{a,b}(i,j-1) + 1 \\
\text{lev}_{a,b}(i-1,j-1) + 1_{(a_i \neq b_j)}
\end{array}
\right. & \text{otherwise}
\end{cases}
\end{equation}

where $a$ and $b$ denote the two strings being compared, and $i$ and $j$ denote the character indices, respectively. The base case $\min(i,j) = 0$ occurs when at least one string is empty; in this case, the number of edits is equal to the length of the other string. Otherwise, the minimum cost among deletion $\text{lev}_{a,b}(i-1,j) + 1$, insertion $\text{lev}_{a,b}(i,j-1) + 1$, and substitution $\text{lev}_{a,b}(i-1,j-1) + 1_{(a_i \neq b_j)}$ is computed recursively.

\begin{table}[!htbp]
    \centering
    \caption{Performance comparison of state-of-the-art vision–language models on the BLPR-D dataset. Model size is reported in billions of parameters.}
    \label{Model_Val_perf}
\begin{tabular}{lccccc}
    \toprule
        \textbf{Model} & \makecell{\textbf{Average}\\ \textbf{Similarity}} & \textbf{Precision} & \textbf{Recall} & \textbf{F1 Score} & \textbf{Size (B)} \\
    \midrule
    Llama 3.2 Vision    & \textbf{0.8184}    & 0.1718      & \textbf{0.8130}    & 0.2835     & 10.7             \\
    Ministral 3& 0.6893    & 0.6951      & 0.6837    & 0.6894     & 8.9              \\
    DeepSeek-OCR & 0.5312    & 0.0031      & 0.6183    & 0.0061     & \textbf{3.3}              \\
    \midrule
    \rowcolor{gray!12}
    \textbf{Gemma3}    & 0.8038    & \textbf{0.8258}      & 0.7958    & \textbf{0.8106}     & 4.3              \\
    \bottomrule
\end{tabular}
\end{table}

\begin{table}[!htbp]
\small
\centering
\caption{Ablation configurations of the proposed BLPR pipeline. Each row represents a variant obtained by selectively enabling or disabling geometric rectification, illumination correction, and the VLM-based fallback, while maintaining the core detection and OCR modules.}
\label{tab:ablation_pipeline}
\renewcommand{\arraystretch}{1.1}
\setlength{\tabcolsep}{3pt}
\begin{tabular}{lcccc}
\toprule
\textbf{Pipeline} & \makecell{\textbf{Geometric}\\\textbf{Rectification}} & \makecell{\textbf{Illumination}\\\textbf{Enhancement}} & \textbf{VLM} & \makecell{\textbf{YOLO} \\\textbf{Detectors}} \\
\midrule
Raw &
\cellcolor{NoRed!35} No &
\cellcolor{NoRed!35} No &
\cellcolor{NoRed!35} No &
\cellcolor{YesBlue!35} Yes \\

\makecell[l]{No Illumination\\Enhancement} &
\cellcolor{YesBlue!35} Yes &
\cellcolor{NoRed!35} No &
\cellcolor{NoRed!35} No &
\cellcolor{YesBlue!35} Yes \\

No Rectification &
\cellcolor{NoRed!35} No &
\cellcolor{YesBlue!35} Yes &
\cellcolor{NoRed!35} No &
\cellcolor{YesBlue!35} Yes \\

No VLM &
\cellcolor{YesBlue!35} Yes &
\cellcolor{YesBlue!35} Yes &
\cellcolor{NoRed!35} No &
\cellcolor{YesBlue!35} Yes \\

Raw VLM &
\cellcolor{NoRed!35} No &
\cellcolor{NoRed!35} No &
\cellcolor{YesBlue!35} Yes &
\cellcolor{NoRed!35} No \\

\makecell[l]{Preprocessed\\VLM} &
\cellcolor{YesBlue!35} Yes &
\cellcolor{YesBlue!35} Yes &
\cellcolor{YesBlue!35} Yes &
\cellcolor{NoRed!35} No \\
\bottomrule
\end{tabular}
\end{table}

\begin{table}[!htbp]
\centering
\caption{Ablation study results on the BLPR-D dataset using the configurations defined in Table \ref{tab:ablation_pipeline}. The results highlight the contribution of each module and the trade-off between accuracy and inference time.}
\label{tab:pipeline_ablation_metrics}
\begin{tabular}{lccccc}
\toprule
\textbf{Pipeline}& \textbf{Average Similarity}& \textbf{Precision} &\textbf{Recall} &\textbf{F1 Score} &\textbf{Time (ms)} \\
\midrule
Raw               & 0.7246 & 0.9275 & 0.7114 & 0.8052 & \textbf{18} \\
No Illumination Enhancement & 0.7267 & \textbf{0.9303} & 0.7132 & 0.8074 & 33 \\
No Rectification   & 0.7248 & 0.9249 & 0.7114 & 0.8042 & 19 \\
No VLM             & 0.7261 & 0.9273 & 0.7129 & 0.8061 & 34 \\
Raw VLM            & 0.8809 & 0.9001 & 0.8751 & 0.8874 & 1,205 \\
Preprocessed VLM   & 0.8781 & 0.8967 & 0.8720 & 0.8842 & 1,067 \\
\midrule
\rowcolor{gray!12}
\textbf{BLPR (Ours)}  & \textbf{0.8960} & 0.9102 & \textbf{0.8943} & \textbf{0.9022} & 402 \\     
\bottomrule
\end{tabular}
\end{table}

\begin{figure}[!htbp]
    \centering
    \includegraphics[width=\linewidth]{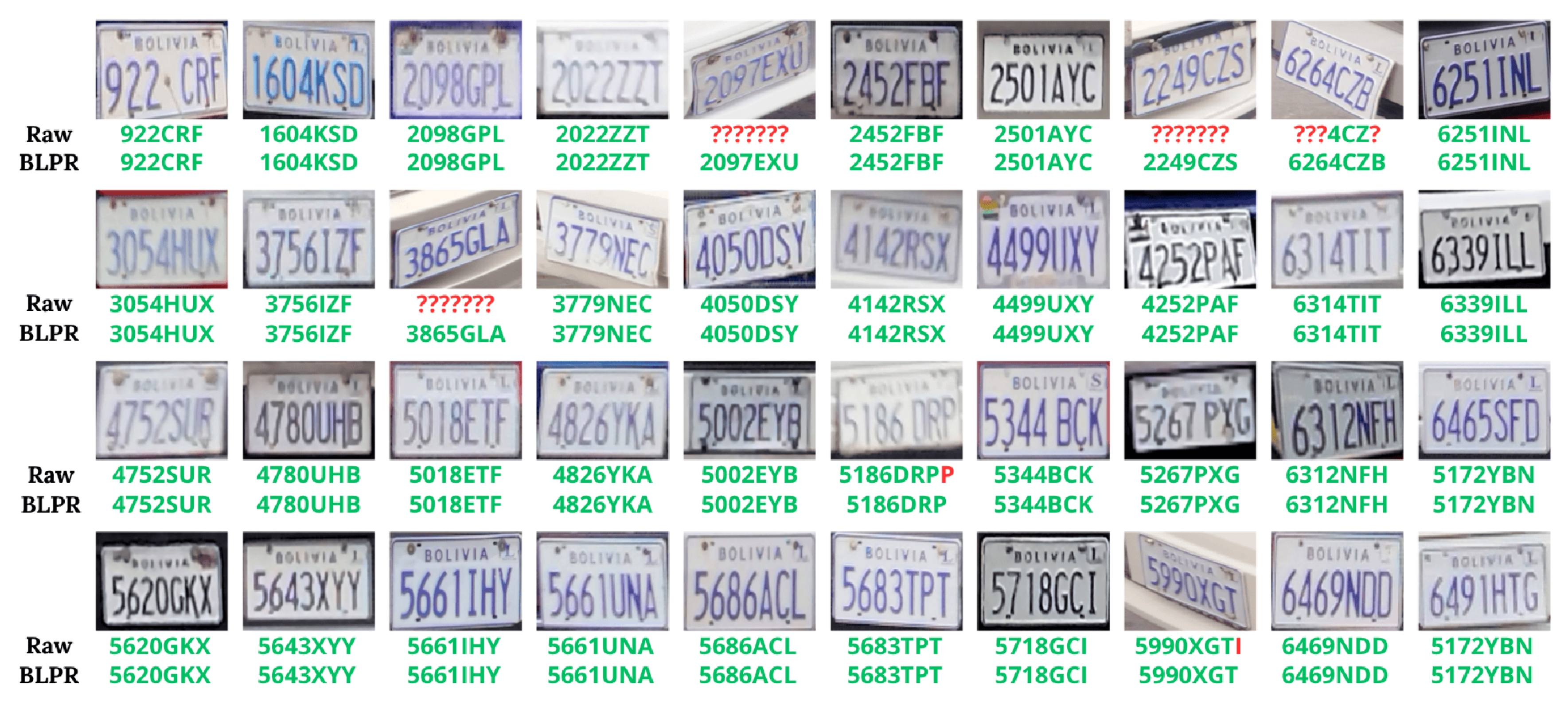}
    \caption{Qualitative results on challenging license plate samples from the BLPR-D dataset, comparing a raw detector-only OCR pipeline (without VLM fallback, geometric rectification, or illumination enhancement modules) with the proposed BLPR system. Correct characters are highlighted in green, whereas errors, including misclassified, hallucinated, or missed characters (denoted by \texttt{?}) are highlighted in red.}
    \label{fig:ocrvsllm_comparison}
\end{figure}

While Llama 3.2 Vision achieved the highest average similarity and recall (0.8184 and 0.8130, respectively), its precision dropped to 0.1718, resulting in an F1-score of 0.2835 and severely limiting its practical utility. Similarly, DeepSeek-OCR and Ministral 3 struggled to maintain a viable balance between precision and recall, yielding F1-scores of 0.0061 and 0.6894, respectively. Gemma3 emerged as the preferred choice, delivering the highest overall F1-score and precision (0.8106 and 0.8258, respectively), while maintaining a competitive average similarity of 0.8038. Its strong performance, combined with a compact architectural footprint, makes it an efficient recognition engine for the proposed system. Crucially, instances in which all evaluated models failed to produce an output were attributed to upstream errors in the car and license plate detection module, which failed to isolate the ROI. These cases were excluded from the evaluation for fairness.

\subsection{Ablation Analysis}

The ablation configurations of the proposed BLPR pipeline are summarized in Table \ref{tab:ablation_pipeline}. Each row represents a variant obtained by selectively enabling or disabling key components, including the YOLO-based OCR detector, geometric rectification, illumination correction, and the VLM-based fallback. As detailed in Table \ref{tab:pipeline_ablation_metrics}, we conducted an ablation analysis on the BLPR-D dataset to evaluate the complete pipeline and isolate the contribution of each module. To this end, we leveraged the distance annotations available in BLPR-D (categorized as ``Near'', ``Normal'', and ``Far'') and excluded plates captured at far distances, as they fall outside the scope of this study.

The baseline ``Raw'' configuration, which relies solely on a YOLO26x model for OCR without preprocessing or fallback mechanisms, achieved an F1-score of 0.8052 with an inference time of 18 ms. Disabling individual preprocessing modules (``No Illumination Enhancement'' and ``No Rectification'') or the fallback mechanism (``No VLM'') resulted in only marginal changes, with F1-scores remaining close to 0.806. This suggests that static preprocessing alone is insufficient to address real-world distortions and provides only limited gains when applied through fixed heuristic rules.

Conversely, across the 470 instances of license plates analyzed in the ablation test, relying exclusively on the VLM modules introduced a substantial and impractical inference latency of 1,067 and 1,205 ms per plate. The proposed BLPR system (``Ours'') effectively balances these trade-offs. By routing plates through targeted geometric and photometric corrections and selectively activating the VLM (which triggered as a fallback in 34.9\% of cases), the method achieved the highest overall performance: an F1-score of 0.9022, an average similarity of 0.8960, and a recall of 0.8943, while reducing the average inference time to a manageable 402 ms.

\begin{table}[!htbp]
\centering
\caption{Detailed ablation results on the BLPR-D dataset across different perspective categories (Normal, Tilted, and Steep) and illumination levels (Low, Medium, and High). The results highlight the contribution of each module under varying conditions and demonstrate that the proposed BLPR system achieves the best trade-off between recognition performance and computational efficiency, particularly in challenging scenarios.}
\label{tab:angle_metrics_detailed}
\begin{adjustbox}{max width=\textwidth}
\begin{tabular}{llcccccc}
\toprule
\textbf{Category} & \textbf{Pipeline} & \textbf{Average Similarity} & \textbf{Precision} & \textbf{Recall} & \textbf{F1 Score} & \textbf{Time (ms)} & Fallback Usage\\
\midrule

\multirow{6}{*}{\textbf{Normal}} 
& Raw               & 0.8515 & 0.9328 & 0.8523 & 0.8907 & \textbf{17} \\
& No Rectification & 0.8526 & 0.9296 & 0.8548 & 0.8906 & 18 \\
& No VLM           & 0.8511 & \textbf{0.9337} & 0.8531 & \textbf{0.8915} & 41 \\
& Raw VLM          & 0.8613 & 0.8869 & 0.8589 & 0.8727 & 1,190 \\
& Preprocessed VLM & 0.8566 & 0.8845 & 0.8515 & 0.8677 & 1,077 \\
\rowcolor{gray!12} \cellcolor{white} & \textbf{BLPR (Ours)} & \textbf{0.8810} & 0.9002 & \textbf{0.8830} & \textbf{0.8915} & 193 & 0.139\\
\midrule
\multirow{6}{*}{\textbf{Tilted}} 
& Raw              & 0.9153 & \textbf{0.9513} & 0.9148 & 0.9327 & \textbf{18} \\
& No Rectification & 0.9142 & 0.9502 & 0.9137 & 0.9316 & \textbf{18} \\
& No VLM           & 0.9159 & \textbf{0.9513} & 0.9159 & 0.9333 & 25 \\
& Raw VLM          & 0.9051 & 0.9163 & 0.8978 & 0.9069 & 1,203 \\
& Preprocessed VLM & 0.9049 & 0.9124 & 0.8988 & 0.9056 & 1,061 \\
\rowcolor{gray!12} \cellcolor{white} & \textbf{BLPR (Ours)} & \textbf{0.9398} & 0.9433 & \textbf{0.9393} & \textbf{0.9413} & 180 & 0.148\\
\midrule
\multirow{6}{*}{\textbf{Steep}} 
& Raw              & 0.4303 & 0.8738 & 0.3920 & 0.5412 & \textbf{18} \\
& No Rectification & 0.4304 & 0.8698 & 0.3903 & 0.5388 & 20 \\
& No VLM           & 0.4345 & 0.8711 & 0.3947 & 0.5432 & 34 \\
& Raw VLM          & \textbf{0.8816} & \textbf{0.9005} & \textbf{0.8735} & \textbf{0.8868} & 1,223 \\
& Preprocessed VLM & 0.8788 & 0.8963 & 0.8717 & 0.8838 & 1,062 \\
\rowcolor{gray!12} \cellcolor{white} & \textbf{BLPR (Ours)} & 0.8756 & 0.8927 & 0.8690 & 0.8807 & 809 & 0.741\\
\bottomrule
\multirow{6}{*}{\textbf{Low Light}} 
& Raw               & 0.4207 & 0.8792 & 0.3784 & 0.5291 & \textbf{18} \\
& No Illumination Enhancement  & 0.4218 & 0.8816 & 0.3794 & 0.5305 & 32 \\
& No VLM            & 0.4226 & 0.8771 & 0.3784 & 0.5287 & 35 \\
& Raw VLM           & \textbf{0.8892} & \textbf{0.9089} & 0.8815 & \textbf{0.8950} & 1,226 \\
& Preprocessed VLM  & 0.8888 & 0.9061 & \textbf{0.8825} & 0.8942 & 1,071 \\
\rowcolor{gray!12} \cellcolor{white} & \textbf{BLPR (Ours)} & 0.8861 & 0.9038 & 0.8794 & 0.8915 & 836 & 0.761\\
\midrule
\multirow{6}{*}{\textbf{Medium Light}} 
& Raw               & 0.8731 & \textbf{0.9544} & 0.8721 & 0.9114 & \textbf{18} \\
& No Illumination Enhancement  & 0.8726 & \textbf{0.9544} & 0.8711 & 0.9108 & 36\\
& No VLM            & 0.8714 & 0.9523 & 0.8702 & 0.9094 & 37 \\
& Raw VLM           & 0.8958 & 0.9047 & 0.8924 & 0.8985 & 1,208 \\
& Preprocessed VLM  & 0.8923 & 0.9009 & 0.8895 & 0.8952 & 1,074 \\
\rowcolor{gray!12} \cellcolor{white} & \textbf{BLPR (Ours)} & \textbf{0.9249} & 0.9307 & \textbf{0.9244} & \textbf{0.9276} & 209 & 0.169\\
\midrule
\multirow{6}{*}{\textbf{High Light}} 
& Raw               & 0.8332 & 0.9229 & 0.8320 & 0.8751 & \textbf{17} \\
& No Illumination Enhancement  & 0.8380 & \textbf{0.9281} & 0.8367 & 0.8800 & 30 \\
& No VLM            & 0.8369 & 0.9249 & 0.8375 & 0.8790 & 31 \\
& Raw VLM           & 0.8626 & 0.8896 & 0.8562 & 0.8726 & 1,187 \\
& Preprocessed VLM  & 0.8587 & 0.8860 & 0.8500 & 0.8676 & 1,059 \\
\rowcolor{gray!12} \cellcolor{white} & \textbf{BLPR (Ours)} & \textbf{0.8802} & 0.8981 & \textbf{0.8812} & \textbf{0.8896} & 230 & 0.185\\
\bottomrule
\end{tabular}
\end{adjustbox}
\end{table}

To gain deeper insights into the system's robustness to spatial distortion, the BLPR-D dataset was manually annotated into three distinct viewing angles defined by the plate's orientation relative to the camera plane: ``Normal'' for a nearly frontal perspective, ``Tilted'' for low-angle perspectives toward the left and right, where characters remain clearly identifiable, and ``Steep'' for high-angle perspectives representing severe spatial distortion. The ablation results for each category are reported in Table~\ref{tab:angle_metrics_detailed}. The ``Raw'' configuration demonstrated strong baseline performance on the easier Normal and Tilted categories but suffered a drop in accuracy, yielding an F1-score of 0.5412 when processing the highly distorted Steep plates. By employing the BLPR system, we bridged this gap and substantially improved the F1-score to 0.8807 for Steep plates by dynamically leveraging targeted geometric corrections alongside the VLM fallback. At the same time, the system tied for the highest F1-score of 0.8915 on Normal plates and achieved the highest F1-score of 0.9413 on Tilted plates, while avoiding unnecessary or potentially harmful preprocessing. These results show that while VLM-based approaches achieve the highest accuracy under severe distortions (e.g., Steep and Low-Light conditions), they also incur substantial computational cost. In contrast, the proposed BLPR system achieves comparable performance while significantly reducing inference time through selective VLM activation.

Figure \ref{fig:ocrvsllm_comparison} compares the predictions generated by the BLPR system on the BLPR-D validation dataset. The first row of predicted texts corresponds to YOLO26x-based OCR, while the second row shows the results obtained after incorporating the Gemma3 VLM fallback mechanism. Characters that match the ground truth are highlighted in green, whereas recognition errors, including misclassified, hallucinated, or missed characters denoted by `?', are highlighted in red. This visualization shows how classical object detection and generative AI approaches can complement each other in challenging recognition tasks.

For cross-dataset generalization evaluation of the proposed BLPR system, we compared its performance with several state-of-the-art (SOTA) datasets from neighboring countries. It is important to note that neither the LPD nor the OCR modules were fine-tuned on any of the datasets to evaluate cross-domain performance. Consequently, the models were applied off-the-shelf, having been trained exclusively on a different domain (Bolivian license plates).

As summarized in Table \ref{tab:cross-domain-comparison}, the proposed BLPR framework achieves robust performance on its native domain, achieving an Average Similarity of 0.8960 and a Precision of 0.9102 on the BLPR-D dataset. Conversely, the evaluation reveals a performance drop when processing foreign domain datasets. Across these test sets, Average Similarity stabilizes in the 0.75 to 0.77 range, while Precision scores fall between 0.8097 and 0.8228. This performance degradation is an expected consequence of domain shift. Metric variations stem from license plate layouts, typography standards, and color coding that are distinct from the Bolivian standard. Nevertheless, sustaining a precision above 80\% on unseen, zero-shot domains highlights the underlying robustness of the BLPR architecture by integrating targeted corrections with a dynamic VLM fallback mechanism. Consequently, the framework mitigates the structural and spatial variations encountered in diverse, real-world deployments.

\begin{table}[!htbp]
\centering
\caption{Cross-domain comparison using the test subset from different benchmark datasets.}
\label{tab:cross-domain-comparison}
\begin{tabular}{lcc}
\toprule
\textbf{Dataset} & \textbf{Average Similarity} & \textbf{Precision} \\
\midrule
SSIG-SegPlate\cite{SegPlate2016} & 0.7749 & 0.8139 \\
UFPR-ALPR\cite{UFPR-ALPR2018}  & 0.7693 & 0.8097 \\
RodoSol-ALPR\cite{laroca2022} & 0.7516 & 0.8228\\
\midrule
\rowcolor{gray!12}
\textbf{BLPR-D} & \textbf{0.8960} & \textbf{0.9102}\\
\bottomrule
\end{tabular}
\end{table}

\section{Discussion}

The results indicate that the primary gains of the proposed system arise from the selective activation of the VLM fallback in challenging scenarios, particularly under steep viewpoint distortion and low-light conditions. This suggests that the modular design effectively separates easy and hard cases, allowing efficient models to process most inputs while reserving computationally expensive reasoning for ambiguous samples. As evidenced by the ablation results, the proposed strategy achieves a favorable balance between accuracy and efficiency by activating the VLM only when necessary, rather than relying on it as a primary recognition mechanism.

The ablation analysis further reveals that static preprocessing techniques, when applied uniformly, provide only marginal improvements across diverse conditions. In contrast, the combination of targeted geometric rectification, illumination correction, and conditional VLM fallback yields consistent and substantial gains. This highlights the importance of adaptive processing strategies that respond to the specific characteristics of each input, rather than applying fixed transformations across all samples. However, these improvements come at the cost of increased inference time when the fallback mechanism is triggered, reflecting an inherent trade-off between recognition accuracy and computational efficiency.

Despite its strong performance, the proposed BLPR system relies on empirically selected heuristic parameters, such as tilt thresholds and foreshortening ratios for license plate enhancement. While these heuristics are effective in practice, they may limit generalization to unseen conditions. Future work should therefore focus on collecting more diverse and representative data, as well as exploring end-to-end learning-based approaches for rectification and OCR that can automatically adapt to complex real-world scenarios. This is particularly important for handling previously unseen license plate formats and enabling reliable deployment in mobile monitoring settings, such as in-vehicle camera systems.

In addition, the current study focuses on license plates captured at moderate distances ($\leq 10$ meters). Extending the BLPR dataset to include far-distance scenarios would enable the development of specialized preprocessing techniques for recovering low-resolution textual information, which remains a key limitation in real-world deployments. In such cases, license plates often occupy only a small fraction of the image, leading to degraded character visibility, increased noise, and higher sensitivity to motion blur and compression artifacts. To address these challenges, future work should explore super-resolution methods, scale-aware detection strategies, and multi-frame or temporal aggregation techniques to enhance recognition robustness in video-based settings.

Finally, the integration of VLMs within the proposed framework presents several opportunities for improvement. Domain-specific fine-tuning of lightweight VLMs, tailored to the structural and typographic constraints of license plates, could enhance both recognition accuracy and inference efficiency. Moreover, optimizing these models for edge deployment through techniques such as quantization and structured pruning is a promising direction, particularly for mobile and embedded applications where latency and energy constraints are critical.

Despite these advances, the proposed system may fail in cases of severe motion blur or when the detection module fails to localize the license plate, highlighting the dependency of downstream modules on ROI quality. This limitation underscores the importance of improving detection robustness as a key component of future end-to-end LPDR systems.

\section{Conclusions}
\label{sec:conclusions}

This work presents a robust license plate detection and recognition system, named BLPR, tailored to the Bolivian context. The modular framework achieves precision and F1-scores exceeding 0.90, while attaining a Levenshtein similarity close to 0.90, demonstrating strong character-level accuracy. By combining a fast OCR backbone with a conditional VLM fallback (Gemma3), the system effectively balances accuracy and computational efficiency, avoiding the latency associated with fully generative AI-based approaches. Furthermore, a key contribution of this work is the preprocessing module guided by lightweight, rule-based heuristics. Rather than applying geometric and photometric corrections uniformly, the system selectively activates these transformations based on measured distortions, thereby mitigating the risk of introducing artifacts such as interpolation blur or contrast degradation and preserving the integrity of the recognition stage.

Finally, this work provides a comprehensive set of resources to support future LPDR research. The BLPR dataset suite (A–D), combined with the BLPR system, establishes a reproducible and extensible framework for developing and evaluating license plate recognition systems in underrepresented environments. Overall, the proposed framework is not limited to the Bolivian context and can be extended to other license plate recognition scenarios with similar constraints, particularly in low-resource and high-variation environments.

\section*{Declarations}
\subsection*{Competing Interests}
The authors declare no conflict of interest.

\subsection*{Code and Data Availability}

The implementation, dataset access instructions, trained-model information, and demo materials are available at: \url{https://github.com/EdwinTSalcedo/BLPR}. The BLPR dataset is publicly available for academic research through a controlled-access process described in the repository.

\subsection*{Author Contributions}
Conceptualization, G.A., D.C., N.C., E.S., S.C.; methodology, G.A., D.C., N.C., S.C.; software, G.A., D.C., N.C.; validation, G.A., D.C., N.C.; investigation, G.A., D.C., N.C., E.S., S.C.; data curation, G.A., D.C., N.C., E.S., S.C.; writing---original draft preparation, G.A., D.C., N.C., E.S., S.C.; writing---review and editing, E.S.; visualization, G.A., D.C., N.C., E.S., S.C.; supervision, E.S.; project administration, E.S. All authors have read and approved this version of the manuscript.

\bibliographystyle{bst/sn-mathphys-num}
\bibliography{bibliography}

\newpage
\appendix
\section*{Appendix}
\addcontentsline{toc}{section}{Appendix}

\section{Geometric Rectification Algorithm}
\label{anx:preprocessing-algorithms}

\begin{algorithm}[!htbp]
\small
\caption{Geometric Rectification}\label{alg:rectification}
\begin{algorithmic}[1]
\Require $ROI$.
\Ensure Rectified $ROI$
\Procedure{GeometricRectification}{$ROI$}
    \State $Polygon \gets \text{ExtractLargestQuadrilateral}(ROI)$ \Comment{Via CLAHE, Canny, and ApproxPolyDP}
    
    \If{$Polygon \neq \textbf{null} \textbf{ and } \text{Area}(Polygon) > 0.15 \cdot \text{Area}(ROI)$}
        \State $FR, \theta_{tilt} \gets \text{CalculateGeometry}(Polygon)$ \Comment{Foreshortening Ratio and Angle}
        
        \Statex \hfill \textit{// Route 1: Severe Distortion}
        \If{$FR > 1.05 \textbf{ or } \theta_{tilt} > 10.0^{\circ}$}
            \State $H, \Delta_{max} \gets \text{CalcHomography}(Polygon)$
            \If{$\Delta_{max} \le 0.20 \cdot \text{Width}(ROI)$}
                \State \Return $\text{WarpPerspective}(ROI, H)$
            \EndIf
        \EndIf

        \Statex \hfill \textit{// Route 2: Flat Plates}
        \If{$FR < 1.05 \textbf{ and } \theta_{tilt} < 5.0^{\circ}$}
            \State $Blob \gets \text{ExtractTextBlob}(ROI)$ \Comment{Denoising and morphological closing}
            \If{$\text{Solidity}(Blob) > 0.45 \textbf{ and } \text{ValidArea}(Blob)$}
                \State $H_{gentle} \gets \text{CalculateHomography}(Blob)$
                \State \Return $\text{WarpPerspective}(\text{Denoise}(ROI), H_{gentle})$
            \EndIf
        \EndIf
    \EndIf

    \Statex \hfill \textit{// Route 3: Moderate Tilt or Guardrail}
    \State \Return $ROI$ \Comment{Pass-through avoids interpolation blur}
\EndProcedure
\end{algorithmic}
\end{algorithm}

\section{Illumination Correction Algorithm}
\label{anx:illumination-correction}

\begin{algorithm}[!htbp]
\color{black}
\small
\caption{Illumination Correction}\label{alg:illumination}
\begin{algorithmic}[1]
\Require $ROI$.
\Ensure Corrected $ROI$
\Procedure{PhotometricCorrection}{$ROI$}
    \State $\mu_v, \sigma_v \gets \text{LuminanceMeanAndStd}(ROI)$ \Comment{Extract Value channel stats}
    
    \Statex
    \If{$\sigma_v > 50 \textbf{ or } (90 \le \mu_v \le 170)$}
        \State \Return $ROI$ \Comment{YOLO visibility confirmed; abort}
    \EndIf

    \Statex
    \State $\gamma \gets \frac{\log(128/255)}{\log(\mu_v/255)}$ \Comment{Dynamic Gamma multiplier}
    \State $\gamma_{clamp} \gets \max(0.6, \min(1.5, \gamma))$ \Comment{Guardrail: prevent overcorrection}
    
    \State $ROI_{\mathrm{final}} \gets \text{ApplyGammaCorrection}(ROI, \gamma_{\mathrm{clamp}})$
\State \Return $ROI_{\mathrm{final}}$
\EndProcedure
\end{algorithmic}
\end{algorithm}

\end{document}